\pdfoutput=1

\documentclass[11pt]{article}

\usepackage[final]{acl}

\usepackage{times}
\usepackage{latexsym}
\usepackage{amsmath}
\usepackage{graphicx}
\usepackage{amssymb}
\usepackage{multirow}
\usepackage{booktabs}
\usepackage[most]{tcolorbox}
\usepackage{paralist}
\usepackage[table]{xcolor}

\usepackage{lineno}

\definecolor{darkblue}{rgb}{0, 0, 0.5}
\hypersetup{colorlinks=true, citecolor=darkblue, linkcolor=darkblue, urlcolor=darkblue}

\usepackage[T1]{fontenc}

\usepackage[utf8]{inputenc}

\usepackage{microtype}

\usepackage{inconsolata}

\usepackage{graphicx}

\title{To Generate or Discriminate? Methodological Considerations for Measuring Cultural Alignment in LLMs}

\author{Saurabh Kumar Pandey\textsuperscript{\thanks{Both authors contributed equally to this paper.}}, Sougata Saha\textsuperscript{\footnotemark[1]}, Monojit Choudhury\\
Mohamed bin Zayed University of Artificial Intelligence\\ 
\texttt{saurabh2000.iitkgp@gmail.com, \{sougata.saha, monojit.choudhury\}@mbzuai.ac.ae}\\
}

\begin{document}
\maketitle
\begin{abstract}

Socio-demographic prompting (SDP) - prompting Large Language Models (LLMs) using demographic proxies to generate culturally aligned outputs - often shows LLM responses as stereotypical and biased. While effective in assessing LLMs' cultural competency, SDP is prone to confounding factors such as prompt sensitivity, decoding parameters, and the inherent difficulty of generation over discrimination tasks due to larger output spaces. These factors complicate interpretation, making it difficult to determine if the poor performance is due to bias or the task design. To address this, we use inverse socio-demographic prompting (ISDP), where we prompt LLMs to discriminate and predict the demographic proxy from actual and simulated user behavior from different users. We use the Goodreads-CSI dataset \citep{saha2025reading}, which captures difficulty in understanding English book reviews for users from India, Mexico, and the USA, and test four LLMs: Aya-23, Gemma-2, GPT-4o, and LLaMA-3.1 with ISDP. Results show that models perform better with actual behaviors than simulated ones, contrary to what SDP suggests. However, performance with both behavior types diminishes and becomes nearly equal at the individual level, indicating limits to personalization.

\end{abstract}

\section{Introduction}

\begin{quote}
    \textit{``It is the novel bits of behaviour, the acts that couldn't plausibly be accounted for in terms of prior conditioning or training or habit, that speak eloquently of intelligence; but if their very novelty and unrepeatability make them anecdotal and, hence, inadmissible evidence, how can one proceed to develop the cognitive case for the intelligence of one's target species?''} - \citet{dennett1988intentional}
\end{quote}

\noindent
Human behavior, broadly defined as the preference for certain values, artifacts, knowledge, etc \citep{Hogg_2016}, is inherently complex and highly variable among individuals \cite{brunswik1955representative, henrich2010weirdest, markus2014culture}. However, patterns do emerge in certain aspects when behavior is aggregated among user groups, which are loosely defined by a \textit{demographic proxy}~\citep{adilazuarda-etal-2024-towards} such as combinations of country, religion, etc. Such patterns constitute a proxy's \textit{prototypical} behavior, which are the most typical or frequent behaviors, values, or norms observed in the group, and also allow for variation and exceptions \cite{rosch1975cognitive, holland1987cultural}. Nonetheless, assessments of LLMs' cultural biases \citep{10.1145/3442188.3445922, Masoud2023CulturalAI} often reduce behavior to \textit{stereotypes}, which are grossly oversimplified and often exaggerated beliefs about the traits or behaviors of members of a demographic proxy \cite{tajfel1979individuals, lippmann2017public}. Probing models using SDP \citep{li2024culture, alkhamissi-etal-2024-investigating, wan-etal-2023-personalized}, such studies usually test for specific sociocultural knowledge through specially curated datasets \citep{nguyen2023extracting, dwivedi-etal-2023-eticor, fung2024massively}. They assume that certain knowledge is central to and therefore known to most members of a demographic proxy, which the models must therefore know \citep{nguyen2023extracting, nguyen2024cultural, shen2024understanding, naous2023having, kotek2023gender, shrawgi2024uncovering}, which is faulty and does not optimally measure models' cultural awareness \cite{saha2025meta, zhou2025culture}. Several studies have also highlighted SDP's prompt sensitivity, which could lead to potentially misleading results \citep{mukherjee-etal-2024-cultural,beck-etal-2024-sensitivity}. Also, computationally, SDP is inherently complex as the output space of generative tasks is usually large \cite{ng2001discriminative, li2008introduction}.

We argue that apart from knowing a culture's \textit{stereotypical behavior}, a model's cultural awareness should also encompass the \textit{broader understanding} and knowledge of the \textit{prototypical behavior}. However, since SDP is constrained by assumptions and limiting factors, it becomes difficult to measure the broader form of cultural awareness. As a solution, we propose ISDP, where we reverse the SDP task by providing the user behavior and asking the model to guess the probable membership of the user across different demographic proxies. We hypothesize that \textit{if indeed models only understand stereotypes, then they should be able to guess the demographic proxy better from simulated behaviors rather than actual user behaviors}. 

It is crucial to distinguish stereotypical behavior in SDP versus ISDP setups. In the ISDP context, a stereotypical behavior can be understood as the model either always associating a particular behavior (text span) with a culture(s) or never associating it, regardless of the surrounding context or user history. Whereas, in SDP, a model's stereotypical behavior would translate to systematically generating the same behavior (response) irrespective of the context.

We test our hypothesis on the Goodreads-CSI dataset \cite{saha2025reading}, which captures incomprehensibility of English book reviews by users from the USA, Mexico, and India, where country is the demographic proxy and incomprehensibility is the behavior. 
We use SDP to simulate user behavior with four LLMs: Aya, Gemma, GPT-4o, and Llama, and then use them as discriminators in ISDP. To our surprise, our hypothesis turned out to be partially false, where GPT-4o is better at predicting the country from actual user behavior rather than simulated behavior. On the contrary, other LLMs perform better with simulated behavior, except for a few instances.

\section{Methodology}

\begin{figure*}[ht!] 
    \centering
    \includegraphics[width=.8\textwidth]{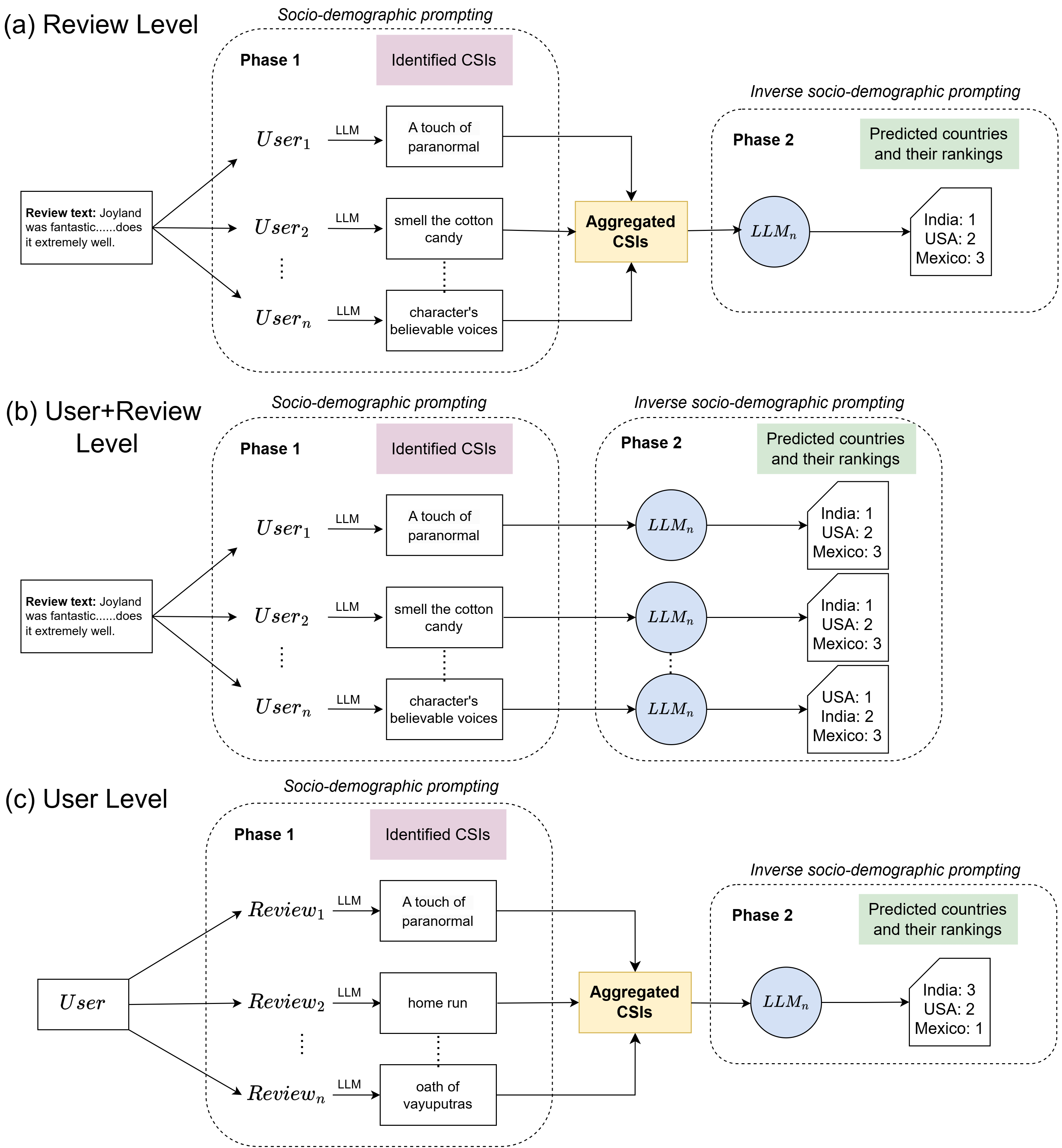}
    \caption{Experimental setup for all three levels of behavior aggregation. Phase 1: SDP and Phase 2: ISDP.}
    \label{fig:overall_flow}
\end{figure*}

Figure~\ref{fig:overall_flow} illustrates our study's overall setup. We discuss the details in this section.

\subsection{Dataset}
\label{sec:experimental_setup}

\textbf{User Behavior}: We use the Goodreads-CSI dataset \citep{saha2025reading}, which contains Culture-Specific Item (CSI) annotations of 57 Indian, Ethiopian, and US English book reviews by 50 users, from India, Mexico, and the USA. CSIs \cite{aixela1996culture} are difficult to understand spans that people agreeably do not understand from a culture, where the incomprehensibility can be attributed to their distinct cultural background. 

\noindent
\textbf{Simulated Behavior}: Similar to \citet{saha2025reading}, we simulate user behavior on the dataset using SDP, where an LLM is tasked to assume the role of a cultural reading assistant. Defining a user at the intersection of two proxies - {\em country} and {\em genre preference}, we use the prompt in Appendix~\ref{appendix:prompts_sdp} to simulate behavior using GPT-4o, Aya-23, Gemma-2, and Llama-3.1. 

\begin{figure}[h!] 
    \centering
    \includegraphics[width=\columnwidth]{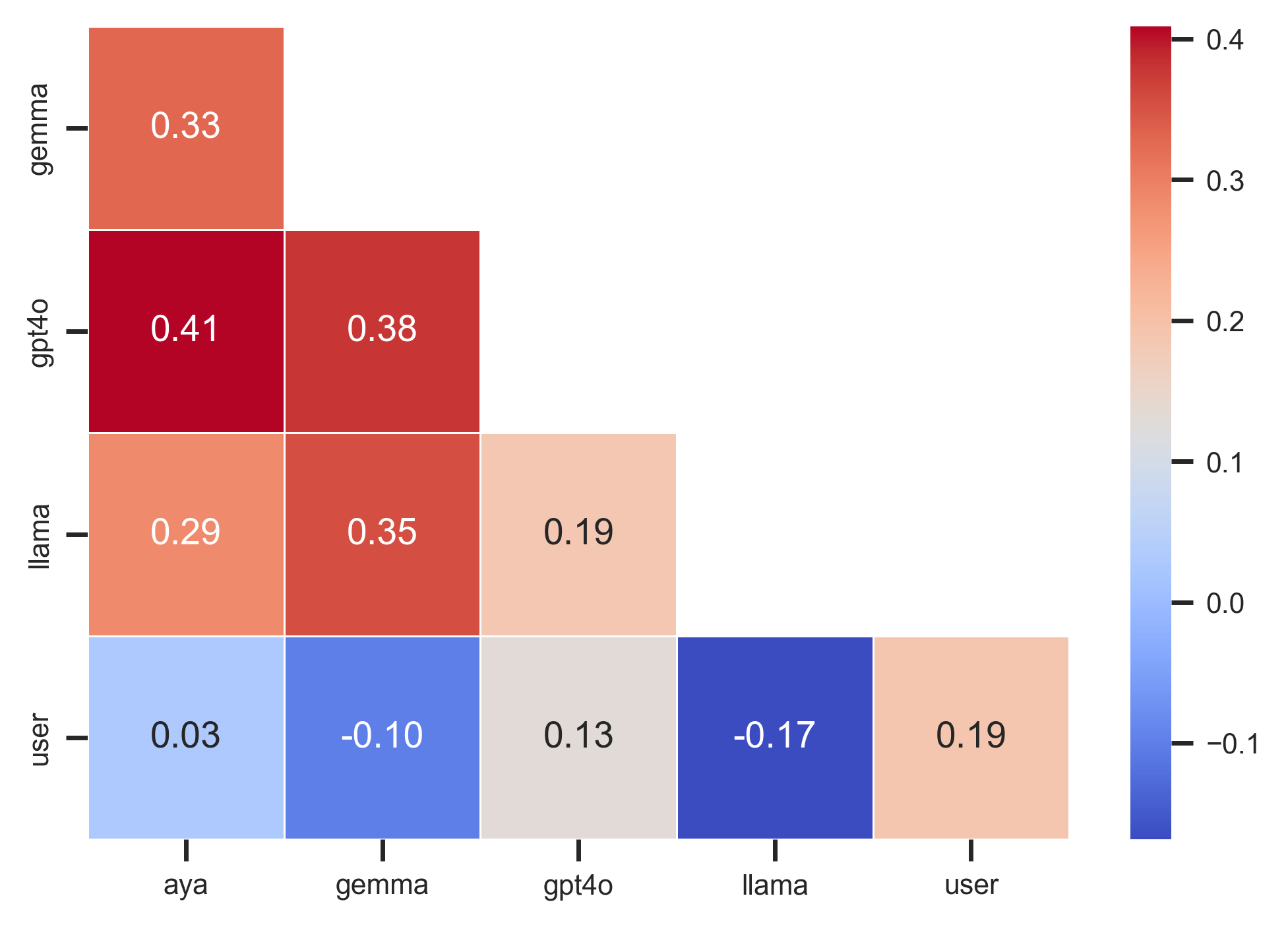}
    \caption{IAA between models and humans for SDP.}
    \label{fig:iaa_heatmap}
\end{figure}

Figure \ref{fig:iaa_heatmap} depicts the Inter Annotator Agreement (IAA) calculated using Krippendorff's alpha between models and users. We observe that the user-user and model-model IAA scores are significantly higher than the user-model scores, implying that models and users agree more among themselves on what is a CSI and exhibit similar behaviors among themselves rather than with each other. The model-model IAA values are also higher in comparison to the user-user IAA, which indicates that although models generate similar (possibly stereotypical) behavior for a demographic proxy, the actual user behavior is much varied. For example, irrespective of user history, all models consistently flagged the span ``FDA'' (Food and Drug Administration, USA) as a CSI, even for users whose histories suggested familiarity with American culture. Conversely, in cases where user histories indicated unfamiliarity with American culture, models sometimes overlooked spans like ``home run'' (a baseball reference), which should have been identified as a CSI.

\noindent
\textbf{Behavior Groups}: We also aggregate the user behavior (original and simulated) at three different levels, to capture a model's ISDP capacity with different amounts of behavior: \textbf{(i) Review level:} Aggregates all CSIs (user behaviors) for a review at a country level. \textbf{(ii) User + Review level:} Default level pertaining to individual user behavior for each review for each country. \textbf{(iii) User level:} Aggregates all CSIs across all reviews for a user.

\subsection{Method}
\textbf{Hypothesis}: We test the hypothesis that given two models, $M_1$ and $M_2$, if $M_1$ is tasked to simulate the behavior of a user from a group $g$ (in this case, country) using SDP, and $M_2$ is tasked to guess $g$ from this simulated behavior $b_{sim}$ using ISDP, then models will be able to predict $g$ better from $b_{sim}$ rather than from real user's behavior $b$. This hypotheses arise from the observation that models trained on similar datasets are likely to generalize in comparable ways and will have similar biases.

\noindent
\textbf{Models}: We experiment with \texttt{Aya-23-8B} \citep{aryabumi2024aya}, \texttt{Gemma-2-9B-it} \citep{team2024gemma}, \texttt{Llama-3.1-8B-Instruct} \citep{dubey2024llama}, and \texttt{gpt-4o-2024-05-01-preview} \citep{achiam2023gpt}, setting temperature to 0.

\noindent
\textbf{Task}: Given a set of behavior ($s \in \{b, b_{sim}\}$), we prompt the models to rank three countries $c \in \{\text{India, Mexico, USA}\}$, according to how likely is the behavior from an individual or a group of users from each country: $Rank(P(g=c|s))$. Each prompt is repeated 5 times, where the ordering of the countries is randomized in each prompt to account for any positional bias. The country with rank 1 indicates individuals or groups from that country would most likely not understand the CSI spans. The country ranked 3 is where the CSIs are least likely to be misunderstood. Performance is evaluated using the Mean Reciprocal Rank (MRR). 

\section{Results and Analysis}
\begin{figure*}[t] 
    \centering
    \includegraphics[width=.95\textwidth]{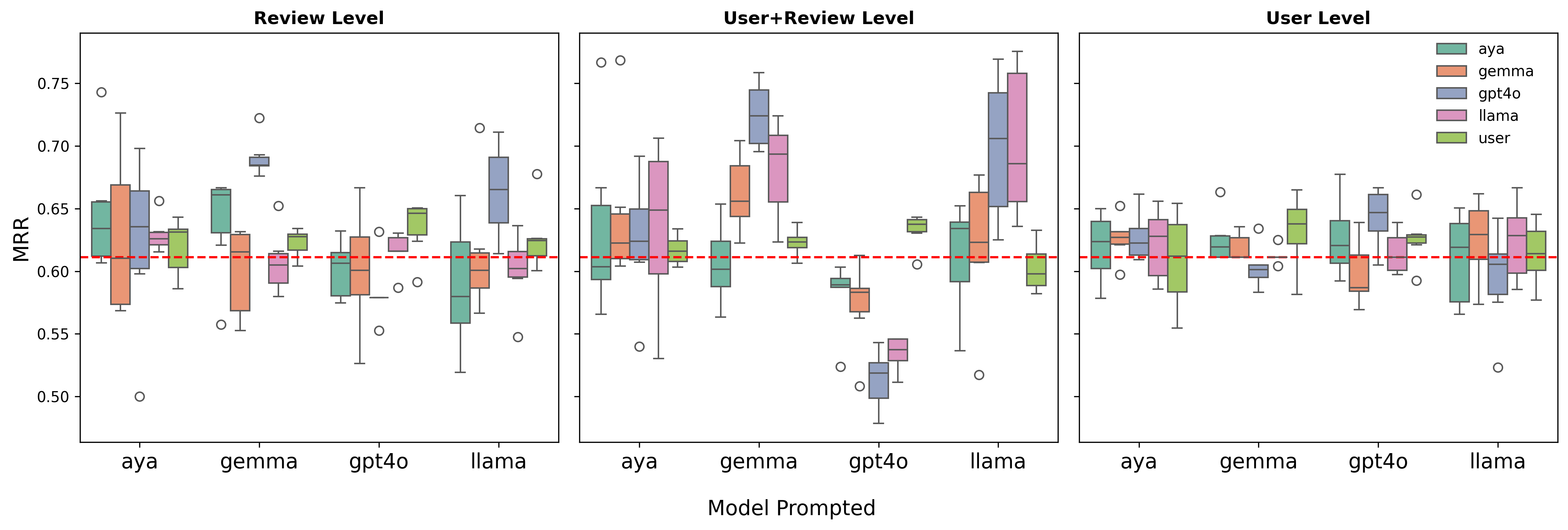}
    \caption{Model-wise MRR scores across five different sources of behavior and three levels of behavior aggregation. The x-axis represents the discriminators, whereas the legend represents the generators used in the experiments.}
    \label{fig:results_boxplot}
\end{figure*}

Figure~\ref{fig:results_boxplot} plots the MRR for all models and all levels of behavior aggregation. The spread of each box depicts the MRR variance due to repeated prompting. The red dotted line (Y=0.61) is the \textit{random baseline}, where each option is likely with a 1/3 probability. Statistical significance of the results using paired t-test is presented in Tables \ref{tab:t_test_review_level}, \ref{tab:t_test_user_review_level}, and \ref{tab:t_test_user_level} in Appendix \ref{sec:ttests}. We observe the following for the ISDP task:

\noindent
\textbf{1.} For \textbf{user-generated spans}, across \textbf{all levels} of behavior aggregation, the average MRR of almost all models is greater than the \textit{random baseline}, indicating ``understanding'' of prototypical behavior to an extent. GPT-4o has the highest average MRR at the \textbf{Review and User+Review} levels, whereas Gemma has the highest score at the \textbf{User} level, for such spans. Interestingly, for \textbf{Review and User+Review} levels, GPT-4o consistently performs worse with model-simulated behavior (including itself), compared to user-generated spans, proving our hypothesis wrong. 


\noindent
\textbf{2.} Contrasting GPT-4o, Gemma, Llama, and Aya perform better with simulated behavior rather than user-generated behavior. Gemma and Llama consistently perform better with \textbf{GPT-4o-generated spans}; Their scores are further amplified at the \textbf{User+Review} level than the \textbf{Review} level. Although this behavior aligns with our hypothesis, it raises questions about the source and nature of the data they are trained on. Furthermore, since the difference between GPT-4o's scores with user vs. its own generated spans increases at the \textbf{User+Review}, we conjecture that GPT-4o (and other models) possibly generates stereotypes during SDP. 



\noindent
\textbf{3.} At the \textbf{User} level, the average MRR scores across models are less varied and their performance is near baseline. Interestingly, GPT-4o performs best on spans generated by itself, which sharply contrasts the trend observed at \textbf{Review} and \textbf{User+Review} levels. We conjecture that non-GPT models are possibly trained on stereotypical data, which enables their performance with simulated data at the \textbf{Review and User+Review} levels. However, since each individual's behavior across all reviews is aggregated at the \textbf{User} level, even if the review-level behaviors were stereotypical, the aggregate possibly reflects the prototypical behavior better, causing all discriminators to perform poorly.

\begin{figure*}[h!] 
    \centering
    \includegraphics[width=\textwidth]{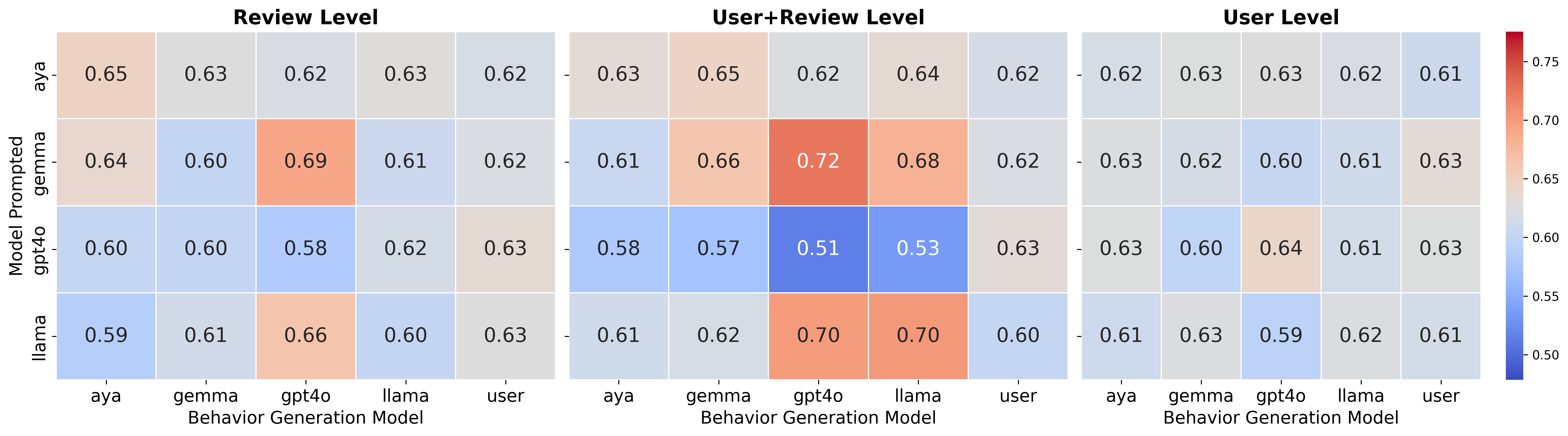}
    \caption{Average MRR scores for different combinations of generators and discriminators.}
    \label{fig:results_heatmap}
\end{figure*}

\noindent
\textbf{4.} Plotting the average MRR scores between different pairs of behavior generators and discriminators in Figure~\ref{fig:results_heatmap} shows our hypothesis does not hold; There are cases, apart from GPT-4o, where models are better discriminators when the behavior is generated by users rather than LLMs.

\noindent
\textbf{5.} We plot the average MRR's trend across the three levels of behavior aggregation for each model in Figure~\ref{fig:results_linecharts}. Similar to the previous findings, the trends indicate that GPT-4o is possibly better at personalization to users, and possibly models user behavior better than other models. This finding aligns with \citet{saha2025user}, where they show the empirical limits of LLMs' personalization capacity \cite{fan2006personalization, lury2019algorithmic} to different sizes of user groups, ranging from a country-level to an individual. Furthermore, this also indicates that other models are essentially learning stereotypes, which further raises questions about the nature of their training data. This might also indicate that the other (smaller) models are trained on GPT or other LLM-generated data, causing them to model user behavior differently than evident in the real world, ascertaining which we leave as future work.

\section{Discussion and Conclusion}
Since LLMs are being used in several everyday applications, strong alignment of their generated behavior to over-represented norms or {\em typical} behavior poses serious risks, not only for the underrepresented communities (say, Mexico or India), but also for users from over-represented groups (say, USA). This is because, as our study shows, every user exhibits some behavior that is not part of the norm. This was demonstrated by \cite{agarwal2024aisuggestionshomogenizewriting} in their study on writing style, where they showed that LLM-assisted writing results in convergence of styles for users from both the USA and India. However, the degree of loss of style diversity was greater for users from India (a further underrepresented group) than for those from the USA. Our study sheds light on why this might be the case, not only for writing styles but for any other aspect of user behavior that LLM-driven applications are expected to replicate or interpret.

A key question remains: \textit{when should we use SDP versus ISDP to measure a model's cultural competency?} Although SDP is widely used to study cultural bias, it is confounded by prompt sensitivity, decoding choices, and the greater difficulty of generation relative to discrimination due to the much larger output space. ISDP mitigates these issues by reframing the task as discrimination: given a behavior, the model selects from a small candidate set of countries (e.g., three in our experiments). Standardizing the input spans further reduces stylistic variance, such as wording, grammar, and other non-content factors. In our experiments, these constraints make ISDP a more reliable indicator of cultural competency. Nevertheless, ISDP is not a replacement for SDP. Rather, it is a complementary method that, alongside SDP, provides a fuller picture of a model's competency. We leave a systematic validation of this complementarity to future work.

\section*{Limitations}

While our study provides critical insights into the cultural competency of LLMs, it has several limitations. First, we rely only on the Goodreads-CSI dataset to test our hypothesis, which might hurt the generalizability of the results. Nonetheless, the results are useful because, as indicated by \citet{saha2025reading}, the dataset contains CSIs pertaining to different cultural dimensions of Newmark's taxonomy \cite{newmark2003textbook}, which cover most aspects of culture. They also justify the dataset's size and argue why it suffices as an evaluation benchmark. Furthermore, the dataset captures knowledge as behavior, which is a unique and cognitively more challenging aspect than other forms of behavior, such as preferences \cite{dunbar1995scientists, kuhl2004early}. Second, we restrict to English reviews while limiting users from only three demographics: India, Mexico, and the USA and which might not be representative of user behavior from all demographics. Finally, all models used in our study were developed in the West. A study incorporating regionally developed models in regional languages would be valuable.

\section*{Acknowledgements}
This research was supported by the Microsoft
Accelerate Foundation Models Research (AFMR) Grant.

\bibliography{acl_latex}

\clearpage
\appendix

\label{sec:appendix}

\section{Appendix}
\begin{figure*}[h!] 
    \centering
    \includegraphics[width=\textwidth]{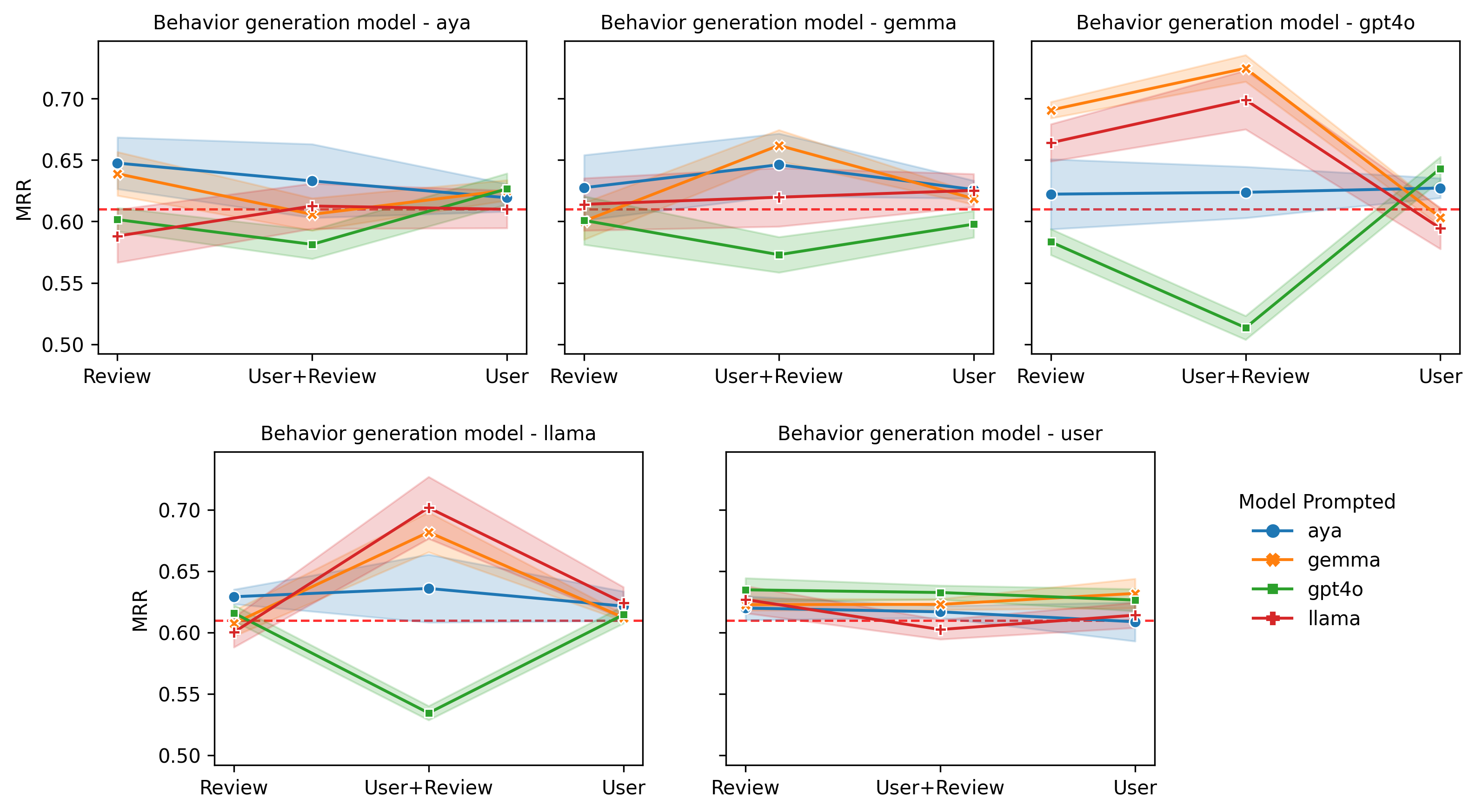}
    \caption{Average MRR scores for different levels of behavior.}
    \label{fig:results_linecharts}
\end{figure*}


\subsection{Prompts - SDP}
\label{appendix:prompts_sdp}

\begin{tcolorbox}[title={Prompt ~- Socio Demographic Prompting}, width=\textwidth, colback=white, colframe=gray, arc=0pt, outer arc=5pt, boxrule=0.5pt, leftrule=2pt, rightrule=2pt, right=0pt, left=0pt, top=0pt, bottom=0pt, toprule=0pt, bottomrule=2pt]
\small

AI Rules\\
- Output response in JSON format\\
- Do not output any extra text. \\
- Do not wrap the outputs in JSON or Python markers\\
- JSON keys and values in double-quotes\\
    
You are a cultural mediator who understands all cultures across the world. As a mediator, your job is to identify and translate culturally exotic concepts from texts from an unknown source culture to my culture. I am a well-educated \{genre\} lover who grew up in \{article\_urban\} urban \{country\}, which defines my culture.
I came across a review of the book '\{book\}' by \{author\}, which belongs to the \{book\_genre\} genre. Given my cultural background, perform the following tasks:\\

Task 1: Identify all culture-specific items (CSIs) from the review text that I might find hard to understand due to my cultural background. CSIs are textual spans denoting concepts and items uncommon and not prevalent in my culture, making them difficult to understand.\\

Task 2:  For each CSI, identify its category from one of the following seven categories:\\
    1. Ecology: Geographical features, flora, fauna, weather conditions, etc.\\
    2. Material: Objects, artifacts, and products specific to a culture, such as food, clothing, houses, and towns.\\
    3. Social: Hierarchies, practices, and rituals specific to a culture.\\
    4. Customs: Political, social, legal, religious, and artistic organizations and practices. Customs, activities, procedures, and concepts.\\
    5. Habits: Gestures, non-verbal communication methods, and everyday habits unique to a culture.\\
    6. Linguistic: Terms unique to a specific language or dialect, including metaphors, idioms, proverbs, humor, sarcasm, slang, and colloquialisms.\\
    7. Other: Anything not belonging to the above six categories.\\

Task 3:  For each CSI, identify its familiarity from one of the following four levels:\\
    1. Familiar: Most people from my culture know and relate to the concept as intended.\\
    2. Somewhat familiar: Only some people from my culture know and relate to the concept as intended.\\
    3. Unfamiliar: Most people from my culture do not know or relate to the concept.\\
    4. Ambiguous: Most people from my culture know the concept, but its interpretation is varied or conflicting.\\

Task 4:  For each CSI, identify its impact on the readability and understandability of the main point of the entire review text from one of the following three levels:\\
    1. High: Greatly hinders the readability and comprehension of the review, making it difficult to convey its main points effectively.\\
    2. Medium: It somewhat affects the readability and comprehension of the review, leading to only partial conveyance of its content.\\
    3. Low: The review text's readability and comprehension will remain unaffected.\\

Task 5: Within 50 words, detail your reason for highlighting the span as CSI in Task 1 by correlating it with my background.\\

Task 6: Explain each CSI span within 20 words to make it more understandable to me. Provide facts, examples, equivalences, analogies, etc, if needed.\\

Task 7: Reformulate the entire text to make it more understandable to me. Keep the length similar to the original review text.\\

Format your response as a valid Python dictionary formatted as: \{'spans': [List of Python dictionaries where each dictionary item is formatted as: \{'CSI': $<$task 1: copy the CSI span from text$>$, 'category': $<$task 2: CSI category name$>$, 'familiarity': $<$task 3:  familiarity level name$>$, 'impact': $<$task 4: impact level name$>$, 'reason': $<$task 5: reason within 50 words$>$, 'explanation': $<$task 6: explain the span within 20 words$>$\}], 'reformulation': $<$task 7: reformulate entire review text$>$\}. Respond with \{'spans': 'None'\} if you think I will not find anything difficult to understand.\\

Text: \{review\_text\}

\end{tcolorbox}

\clearpage

\subsection{Prompts - ISDP}
\label{appendix:prompts_isdp}

\begin{tcolorbox}[title={Prompt ~- Inverse socio-demographic prompting (Review Level/User+Review Level)}, width=\textwidth, colback=white, colframe=gray, arc=0pt, outer arc=5pt, boxrule=0.5pt, leftrule=2pt, rightrule=2pt, right=0pt, left=0pt, top=0pt, bottom=0pt, toprule=0pt, bottomrule=2pt]
\small

AI Rules \\
- Output response strictly in JSON format. \\
- Do not output any extra text or explanations outside the JSON. \\
- Do not wrap the outputs in JSON or Python markers. \\
- JSON keys and values should be enclosed in double quotes. \\
        
You are a cultural mediator who understands all cultures across the world. As a mediator, your job is to identify the cultural background of the users. The user came across a review of the book \{book\} by \{author\}, which belongs to the \{book\_genre\} genre. Culture-specific items (CSIs) are textual spans denoting concepts and items uncommon and not prevalent in a culture, making them difficult to understand for an individual from a given culture. Given the review text and all the CSI spans that the users found hard to understand due to their cultural background, perform the following task. \\

Task 1: Based on the CSI spans that the users found difficult to understand due to their cultural background, provided as a tuple of (CSI span, number of users who found the span difficult to understand), identify the users' likely country of origin. Rank the countries in decreasing order of how unfamiliar the CSI spans are in that country. The possible countries are India, USA, and Mexico. The country ranked highest is the one where the CSIs are least common (most unfamiliar), and the lowest-ranked country is where the CSIs are most common (least unfamiliar). \\

Task 2: Explain within 20 words the reason behind the ranking of the countries in Task 1.\\

Format your response as a valid Python dictionary formatted as: \{'country': $<$task 1: dictionary of all possible countries with their rankings formatted as \{'country': ranking\}$>$, 'reason': $<$task 2: reason within 20 words$>$\}\\

Review Text: \{review\_text\}\\

CSIs: \( \{( \text{span}_1, n_1 ), ( \text{span}_2, n_2 ), \dots, ( \text{span}_N, n_N )\} \)

\end{tcolorbox}

\clearpage
\begin{tcolorbox}[title={Prompt ~- Inverse socio-demographic prompting (User Level)}, width=\textwidth, colback=white, colframe=gray, arc=0pt, outer arc=5pt, boxrule=0.5pt, leftrule=2pt, rightrule=2pt, right=0pt, left=0pt, top=0pt, bottom=0pt, toprule=0pt, bottomrule=2pt]

AI Rules \\
- Output response strictly in JSON format. \\
- Do not output any extra text or explanations outside the JSON. \\
- Do not wrap the outputs in JSON or Python markers. \\
- JSON keys and values should be enclosed in double quotes. \\

You are a cultural mediator who understands all cultures across the world. As a mediator, your job is to identify the cultural background of a user. Culture-specific items (CSIs) are textual spans denoting concepts and items uncommon and not prevalent in a culture, making them difficult to understand for an individual from a given culture. The user came across reviews of books. Given the review text and all the CSI spans that the user found difficult to understand due to their cultural background, perform the following task. \\
    
Task 1: Based on the CSI spans that the users found difficult to understand due to their cultural background, provided as a tuple of (CSI span, number of users who found the span difficult to understand), identify the users' likely country of origin.. Rank the countries in decreasing order of how unfamiliar the CSI spans are in that country. The possible countries are India, USA, and Mexico. The country ranked highest is the one where the CSIs are least common (most unfamiliar), and the lowest-ranked country is where the CSIs are most common (least unfamiliar).\\

Task 2: Explain within 20 words the reason behind the ranking of the countries in Task 1.\\

Format your response as a valid Python dictionary formatted as: \{'country': $<$task 1: dictionary of all possible countries with their rankings formatted as \{'country': ranking\}$>$, 'reason': $<$task 2: reason within 20 words$>$\}\\

Given below is the review of the book \{book\} by \{author\}, which belongs to the \{book\_genre\} genre.
Review Text1: \{review\_text\}\\
CSIs: \( \{( \text{span}_1, n_1 ), ( \text{span}_2, n_2 ), \dots, ( \text{span}_N, n_N )\} \)\\
\\
Given below is the review of the book \{book\} by \{author\}, which belongs to the \{book\_genre\} genre.
Review Text: \{review\_text\}\\
CSIs: \( \{( \text{span}_1, n_1 ), ( \text{span}_2, n_2 ), \dots, ( \text{span}_N, n_N )\} \)\\
.\\
.\\
.\\
.\\
Given below is the review of the book \{book\} by \{author\}, which belongs to the \{book\_genre\} genre.
Review Text: \{review\_text\}\\
CSIs: \( \{( \text{span}_1, n_1 ), ( \text{span}_2, n_2 ), \dots, ( \text{span}_N, n_N )\} \)

\end{tcolorbox}

\subsection{Results: Statistical Significance}
\label{sec:ttests}
\begin{table*}[h]
\centering
\begin{tabular}{l|l|l|r|r}
\toprule
span\_generator & discriminator1 & discriminator2 & t\_statistic & p\_value \\
\midrule
\rowcolor{orange} aya & llama & aya & -5.326117 & 0.003124 \\
aya & llama & gemma & -1.347977 & 0.235514 \\
aya & llama & gpt4o & -0.450854 & 0.670974 \\
aya & aya & gemma & 0.227204 & 0.829263 \\
aya & aya & gpt4o & 1.572355 & 0.176675 \\
\rowcolor{orange} aya & gemma & gpt4o & 3.152963 & 0.025294 \\
gemma & llama & aya & -0.740100 & 0.492505 \\
gemma & llama & gemma & 0.413237 & 0.696558 \\
gemma & llama & gpt4o & 0.467127 & 0.660058 \\
gemma & aya & gemma & 0.769376 & 0.476424 \\
gemma & aya & gpt4o & 0.912667 & 0.403278 \\
gemma & gemma & gpt4o & -0.029062 & 0.977939 \\
gpt4o & llama & aya & 1.083931 & 0.327873 \\
gpt4o & llama & gemma & -1.355720 & 0.233205 \\
\rowcolor{orange} gpt4o & llama & gpt4o & 4.574447 & 0.005978 \\
gpt4o & aya & gemma & -2.422944 & 0.059899 \\
gpt4o & aya & gpt4o & 1.034381 & 0.348370 \\
\rowcolor{orange} gpt4o & gemma & gpt4o & 8.260590 & 0.000424 \\
user & llama & aya & 0.354713 & 0.737275 \\
user & llama & gemma & 0.343953 & 0.744874 \\
user & llama & gpt4o & -0.405212 & 0.702077 \\
user & aya & gemma & -0.251829 & 0.811197 \\
user & aya & gpt4o & -1.647885 & 0.160292 \\
user & gemma & gpt4o & -1.140894 & 0.305595 \\
\rowcolor{orange} llama & llama & aya & -3.120119 & 0.026249 \\
llama & llama & gemma & -0.835801 & 0.441365 \\
llama & llama & gpt4o & -1.177835 & 0.291865 \\
\rowcolor{orange} llama & aya & gemma & 4.332499 & 0.007482 \\
llama & aya & gpt4o & 1.412226 & 0.216988 \\
llama & gemma & gpt4o & -0.618236 & 0.563491 \\
\bottomrule
\end{tabular}
\caption{Paired t-test results for review level experiments. Rows colored orange show that they are statistically significant (p\_value $<$ 0.05).}
\label{tab:t_test_review_level}
\end{table*}

\begin{table*}[h]
\centering
\begin{tabular}{l|l|l|r|r}
\toprule
span\_generator & discriminator1 & discriminator2 & t\_statistic & p\_value \\
\midrule
aya & llama & aya & -0.512273 & 0.630278 \\
aya & llama & gemma & 0.355522 & 0.736705 \\
aya & llama & gpt4o & 1.397820 & 0.221018 \\
aya & aya & gemma & 0.648904 & 0.545023 \\
aya & aya & gpt4o & 1.443309 & 0.208528 \\
aya & gemma & gpt4o & 1.843266 & 0.124623 \\
gemma & llama & aya & -0.569640 & 0.593560 \\
gemma & llama & gemma & -1.727200 & 0.144713 \\
gemma & llama & gpt4o & 1.799847 & 0.131783 \\
gemma & aya & gemma & -0.486936 & 0.646898 \\
\rowcolor{orange} gemma & aya & gpt4o & 2.736133 & 0.040980 \\
\rowcolor{orange} gemma & gemma & gpt4o & 6.344467 & 0.001436 \\
\rowcolor{orange} gpt4o & llama & aya & 3.633479 & 0.015006 \\
gpt4o & llama & gemma & -0.968320 & 0.377358 \\
\rowcolor{orange} gpt4o & llama & gpt4o & 8.703857 & 0.000331 \\
\rowcolor{orange} gpt4o & aya & gemma & -4.606414 & 0.005806 \\
\rowcolor{orange} gpt4o & aya & gpt4o & 5.604685 & 0.002499 \\
\rowcolor{orange} gpt4o & gemma & gpt4o & 12.269834 & 0.000064 \\
user & llama & aya & -1.301253 & 0.249904 \\
\rowcolor{orange} user & llama & gemma & -4.421119 & 0.006885 \\
\rowcolor{orange} user & llama & gpt4o & -3.433467 & 0.018566 \\
user & aya & gemma & -0.779455 & 0.470977 \\
user & aya & gpt4o & -1.842872 & 0.124686 \\
user & gemma & gpt4o & -2.003821 & 0.101443 \\
llama & llama & aya & 2.007264 & 0.100998 \\
llama & llama & gemma & 0.598369 & 0.575666 \\
\rowcolor{orange} llama & llama & gpt4o & 6.860223 & 0.001006 \\
llama & aya & gemma & -1.150194 & 0.302086 \\
\rowcolor{orange} llama & aya & gpt4o & 4.116861 & 0.009202 \\
\rowcolor{orange} llama & gemma & gpt4o & 7.476294 & 0.000676 \\
\bottomrule
\end{tabular}
\caption{Paired t-test results for user+review level experiments. Rows colored orange show that they are statistically significant (p\_value $<$ 0.05).}
\label{tab:t_test_user_review_level}
\end{table*}

\begin{table*}
\centering
\begin{tabular}{l|l|l|r|r}
\toprule
span\_generator & discriminator1 & discriminator2 & t\_statistic & p\_value \\
\midrule
aya & llama & aya & -0.466986 & 0.660152 \\
aya & llama & gemma & -0.800898 & 0.459539 \\
aya & llama & gpt4o & -0.755307 & 0.484104 \\
aya & aya & gemma & -0.679548 & 0.526970 \\
aya & aya & gpt4o & -0.314223 & 0.766040 \\
aya & gemma & gpt4o & -0.063718 & 0.951664 \\
gemma & llama & aya & -0.036038 & 0.972647 \\
gemma & llama & gemma & 0.553196 & 0.603952 \\
gemma & llama & gpt4o & 1.631805 & 0.163649 \\
gemma & aya & gemma & 0.951767 & 0.384923 \\
gemma & aya & gpt4o & 2.026568 & 0.098541 \\
gemma & gemma & gpt4o & 2.157585 & 0.083441 \\
gpt4o & llama & aya & -2.018184 & 0.099601 \\
gpt4o & llama & gemma & -0.724634 & 0.501154 \\
gpt4o & llama & gpt4o & -2.326693 & 0.067489 \\
gpt4o & aya & gemma & 2.038030 & 0.097112 \\
gpt4o & aya & gpt4o & -1.258016 & 0.263937 \\
\rowcolor{orange} gpt4o & gemma & gpt4o & -3.157931 & 0.025154 \\
user & llama & aya & 0.379708 & 0.719756 \\
user & llama & gemma & -1.288302 & 0.254034 \\
user & llama & gpt4o & -1.117324 & 0.314649 \\
user & aya & gemma & -1.075421 & 0.331318 \\
user & aya & gpt4o & -1.512893 & 0.190718 \\
user & gemma & gpt4o & 0.382935 & 0.717507 \\
llama & llama & aya & 0.117893 & 0.910742 \\
llama & llama & gemma & 0.872386 & 0.422905 \\
llama & llama & gpt4o & 1.005745 & 0.360700 \\
llama & aya & gemma & 0.910761 & 0.404190 \\
llama & aya & gpt4o & 0.502238 & 0.636832 \\
llama & gemma & gpt4o & -0.395285 & 0.708934 \\
\bottomrule
\end{tabular}
\caption{Paired t-test results for user level experiments. Rows colored orange show that they are statistically significant (p\_value $<$ 0.05).}
\label{tab:t_test_user_level}
\end{table*}

\clearpage

\subsection{Related Work}

Predicting and modeling human behavior has always been a challenging task \citep{7435176, cui2016computational, wang2024survey}. Measurement of cultural awareness in LLMs inherently requires modeling and/or prediction of human behavior, which is typically conducted using SDP on specifically curated datasets under culture-specific settings \citep{nguyen2023extracting, dwivedi-etal-2023-eticor, fung2024massively, shi-etal-2024-culturebank, nadeem-etal-2021-stereoset, wan-etal-2023-personalized, jha-etal-2023-seegull, li2024culture, cao-etal-2023-assessing, tanmay2023exploring, rao-etal-2023-ethical, kovavc2023large}. Studies have also evaluated LLMs' knowledge of cultural artifacts such as food, art forms, clothing, and geographical markers \citep{seth-etal-2024-dosa, li2024culturellm, koto-etal-2024-indoculture}. However, many of these methods argue that there is a need for the development of robust evaluation benchmarks that can test the cultural understanding in LLMs \citep{wang-etal-2024-cdeval, rao2024normad, myung2024blend, zhou2024does, putri-etal-2024-llm, mostafazadeh-davani-etal-2024-d3code, wibowo-etal-2024-copal, owen2024komodo, chiu2024culturalbench, liu-etal-2024-multilingual, koto-etal-2024-indoculture}. The Goodreads-CSI dataset, recently introduced by \cite{saha2025reading}, serves as a robust benchmark as they capture CSIs- a term introduced by \citet{aixela1996culture} and further explored in various works \citep{pandey-etal-2025-culturally, zhang-etal-2024-cultural, daghoughi2016analysis, narvaez2014translate, sperber1994cross, trivedi2008translating}, which depicts things that people would not understand due to their culture.

\section{Dataset standardization}
\label{appendix:dataset_standardization}

The dataset contains human annotations of 57 English reviews of books originating from India, UAE, and the USA. The annotations are performed by 50 users, where 8 are from India, 22 are from Mexico, and 20 are from the USA. Given a book review, each user identifies text spans of varying lengths that they found difficult to understand. 

\paragraph{Span standardization: } 
Since users can mark any contiguous text spans as difficult to understand, there is a high degree of variability in the annotations. To handle this, the original dataset \citep{saha2025reading} semantically clustered the spans using sentence transformers\footnote{sentence-transformers/all-MiniLM-L6-v2} and filtered out poor-quality annotations before conducting quantitative and qualitative analysis. However, this approach has several limitations: \textbf{(i) User-User Mismatch:} The lengths of user annotations vary where one user might identify multiple CSIs as a single contiguous span, whereas others might segment the same span into multiple non-contiguous spans. For example, the span {\em from the Beats in On The Road to Ken Kesey's Merry Pranksters} refers to two influential countercultural movements in American literature and history represented by {\em the Beats in On The Road} and {\em Ken Kesey's Merry Pranksters}. Someone unfamiliar with either of them might mark the entire span as difficult to understand, while others might partially mark the spans, signifying familiarity with either {\em the Beat generation} or {\em the pranksters}. Some might non-contiguously highlight both spans to indicate unfamiliarity with both of them. Such fine-level distinctions are lost in sentence transformers-based semantic clustering. \textbf{(ii) User-Model Mismatch:} User-annotated spans were generally longer than those generated by the model, primarily due to differences in word boundary recognition. Users often group multiple CSIs into a single span, whereas the model-generated responses tend to break them down into more atomic CSIs. \textbf{(iii) Model-Model Mismatch:} Unlike \citet{saha2025reading}, since we also evaluate multiple models here, there might be scenarios where the model-highlighted spans do not have a 1:1 match, similar to the user-user mismatches. To alleviate these issues, we collected all the CSI spans annotated by users and models for a given review text and manually standardized the spans for each review text, ensuring consistent discourse segmentation. 

\begin{tcolorbox}[width=\columnwidth, colback=white, colframe=gray, arc=0pt, outer arc=5pt, boxrule=0.5pt, leftrule=2pt, rightrule=2pt, right=0pt, left=0pt, top=0pt, bottom=0pt, toprule=2pt, bottomrule=2pt, fontupper=\small]
\textbf{\underline{\textit{Before Standardization:}}} 

\vspace{2pt}
\textbf{User:} John Muir, Muir woods, Stickeen, The Moral Equivalent of War by William James  \\
\textbf{Model 1:} John Muir? Sure, Muir woods, \\ 
\textbf{Model 2:} John Muir  \\
\textbf{Overlap scores. Model 1:} 1.0  \textbf{Model 2:} 1.0

\vspace{5pt}

\textbf{\underline{\textit{After Standardization:}}} 

\vspace{2pt}
\textbf{User:} John Muir\#Muir woods\#Stickeen\#The Moral Equivalent of War\#William James  \\
\textbf{Model 1:} John Muir\#Muir woods  \\
\textbf{Model 2:} John Muir  \\
\textbf{Overlap scores. Model 1:} 0.4  \textbf{Model 2:} 0.2
\end{tcolorbox}

\noindent
Above is an illustrative example, which shows the overlap scores for two models before and after the standardization process. Before standardization, both models 1 and 2 attain a perfect score using sentence transformers with a similarity threshold of 0.5. However, after standardization, the user span is split into five unique spans depicting different CSIs. Using an exact match, we obtain an overlap score of 0.4 for Model 1 and 0.2 for Model 2. 

\noindent
The dataset contains 1,193 combinations of reviews and CSIs annotated by the users and generated by the models. Three Computer Science and Linguistics experts manually standardized all 1,193 spans in the dataset by converting them to their appropriate elementary discourse units (EDUs). Each annotator annotated 450 spans (avg 19 reviews per annotator), which were randomly sampled and had \textasciitilde50 overlapping spans among them to facilitate calculating inter-annotator agreement (IAA) scores. The annotation guidelines for the standardization of spans are as follows.
\begin{compactitem}
    \item If a span contains multiple CSIs, split it into their elementary discourse units separated by a ``\#'' symbol.
    \item If a span contains only part of a named entity (such as a book title, album title, or proper noun), the span should be expanded to include the full entity.
    \item Correct any grammatical errors and formatting inconsistencies, wherever necessary.
\end{compactitem}
Since each annotation either involved correcting the errors in the span or splitting a span into multiple spans, we use a mean-based IAA metric to calculate agreement between the three expert annotators. We assign a weighted score to the agreements and disagreements while calculating IAA and assign a perfect score if the two annotators agree on an annotation and a score of 0.7 otherwise. After the first round of annotation, we obtain a mean agreement of 0.967. The disagreements were discussed and resolved by an additional round of annotation. We observe a decrease in the average number of words in user spans from 6.31 to 3.30, and from 5.22 to 3.36 in the model spans, indicating better consistency. Post standardization, the total review text and span combinations were 922 (322 from users, 600 from models), compared to 1,193 (365 from users, 828 from models). Overall, the dataset contains 671 unique spans across all review texts, compared to 1,122 spans. The standardization process ensures the reliability of any following span-based analysis, enabling more robust comparisons between humans and models.\footnote{We will release the updated dataset with standardized spans upon acceptance.}

\end{document}